\documentclass[10pt,conference,a4paper]{IEEEtran}
\usepackage{times,amsmath,epsfig}
\usepackage{array}
\usepackage{multirow}
\usepackage{subfigure}
\usepackage{graphicx}
\title{Accelerate CU Partition in HEVC using Large-Scale Convolutional Neural Network}
\author{%
{Chenying Wang{\small $~^{\#1}$}, Li Yu{\small $~^{\#1}$}, Shengwei Wang{\small $~^{\#1}$} } %  Removed for anonymous submission
{}
\vspace{1.6mm}\\
\fontsize{10}{10}\selectfont\itshape
$^{\#}$\,School of Electronic Infomation and Communications, Huazhong University of Science and Technology\\ %  Removed for anonymous submission
\,\\ 
\\
\fontsize{9}{9}\selectfont\ttfamily\upshape
$^{1}$\,\{wangchenying, hustlyu, kadinwang\}@hust.edu.cn\\ % Removed for anonymous submission
\vspace{1.2mm}\\
\fontsize{10}{10}\selectfont\rmfamily\itshape
\,\\ 
\\

\fontsize{9}{9}\selectfont\ttfamily\upshape
\,
}
\begin{document}
\maketitle

% INCLUDES COPYRIGHT NOTICE: one of three copyright notice should be included. Uncomment the appropriate line below, according to the authors affiliation:
\begin{figure}[b]
\parbox{\hsize}{\em
%information about the event:
%IEEE VCIP'14, Dec. 7 - Dec. 10, 2014, Valletta, Malta.

%copyright notice: one of three copyright notices below should be included. Uncomment the appropriate line, according to the authors affiliation:
%000-0-0000-0000-0/00/\$31.00 \ \copyright 2014 IEEE.
%U.S. Government work not protected by U.S. copyright.
%???-?-????-????-?/10/\$??.?? \copyright 2014 Crown.
}\end{figure}

\begin{abstract}
High efficiency video coding (HEVC) suffers high encoding computational complexity, partly attributed to the rate-distortion optimization quad-tree search in CU partition decision.
Therefore, we propose a novel two-stage CU partition decision approach in HEVC intra-mode.
In the proposed approach, CNN-based algorithm is designed to decide CU partition mode precisely in three depths.
In order to alleviate computational complexity further, an auxiliary earl-termination mechanism is also proposed to filter obvious homogeneous CUs out of the subsequent CNN-based algorithm.
Experimental results show that the proposed approach achieves about 37\% encoding time saving on average and insignificant BD-Bitrate rise compared with the original HEVC encoder.
\\[1\baselineskip]
\end{abstract}

% NOTE keywords are not used for conference papers so do not populate them

\section{Introduction}

High efficiency video coding (HEVC) \cite{sullivan2012overview} is the state-of-the-art video coding standards developed by the joint collaborative team on video coding (JCT-VC).
HEVC supports four depth levels of coding units (CUs) at maximum, from $64\times64$ to $8\times8$, also known as CU quad-tree partition.
By contrast with the fixed-size marcoblock supported in H.264 \cite{wiegand2003overview}, this flexible partition structure leads that at the meantime of performance enhancement, CU partition decision becomes one of the most time-consuming modules in HEVC encoder.
Currently, CU partition decision relies on a rate-distortion optimization (RDO) quad-tree search algorithm.
Specifically HEVC encoder recursively in three depths, makes decision whether a CU split or not by comparing if the rate-distortion cost of the CU as a whole is higher than the sum of rate-distortion costs of four splited sub-CUs.
In the light of this condition, so much encoding time is spent on CU partition decision that a faster approach of it is urgently required.

In the last few years, deep learning has successfully demonstrated its impressive performance on many varieties of fields.
Convolutional neural networks (CNNs) specifically are able to extract features from images and then classify them to several types.
Accordingly, an idea comes up that we can take advantages of CNNs to assist many decision modules in HEVC encoder.
CNNs are able to boost the decision modules in HEVC encoding, such as CU partition, intra and inter prediction mode.
For instance, CNNs can assist HEVC encoder to decide CU partition much faster than traditional method.
At the meantime, since CNNs are able to achieve considerable accuracy of CU partition decision, RD performance of encoded video will hardly be aggravated.  

Several CNN-based fast CU partition decision algorithms are emerging.
Liu \textit{et al.} \cite{liu2016cnn} proposed a simple CNN structure to predict CU mode.
Liu's work mainly focuses on pipe-line to adapt hardware implementation, which is tough to be ported to another hardware or platform.
Moreover, Li \textit{et al.} \cite{li2017deep} presented a deep CNN to decide three level CU partition, and in their CNN there are three branches with different sizes and quantities of convolution layers.
On the other hand, none of feasible measure is taken during training process against the imbalance training samples issue.
Xu \textit{et al.} \cite{xu2017reducing} raised a novel CNN structure to decide CU quad-tree  partition over three depths at one time.
Within a $64\times64$ CU, Xu's proposed CNNs utilize redundant neighbor sub-CUs to sub-CU partition decision in second and third depth.

In this paper, we propose a novel two-stage approach in order to accelerate CU partition decision.
The first stage of the proposed approach is a rough early-termination algorithm based on the range of CU luma samples that directly related to CU partition mode.
This early-termination algorithm will be applied before CNN-based stage to filter some evident non-split CUs out of the subsequent CNN in order to tightly restrain computational complexity.
The second stage is CNNs designed and trained to precisely estimate the partition mode of ambiguous CUs that the first stage conveys.
Due to the depth and width of our neural networks, our CNNs are able to accomplish much lower error rate.
In a nutshell, the first stage, or the early-termination algorithm, is a rough but fast method to filter some evident homogeneous CUs, which plays an important part in complexity reduction.
Then in the second stage the CNN-based CU partition algorithm is a final method with higher accuracy, which mainly concentrates on RD performance retention.
As a result, the proposed two-stage approach gains the balance of time saving and RD loss.

\section{Two-Stage CU Partition Approach}

CU partition mode is decided over three depths totally and CUs are determined to split or to non-split at each depth. 
In other word, CU partition decision at each depth is actually a binary classification problem, so we design a CU non-split/split classifier.
The proposed approach measures range of CU luma samples firstly to filter homogeneous CUs to early terminate.
Then CUs passed early-termination filter will be decided by three CNNs whether to split or not in each depth.

\subsection{Early-Termination Algorithm}

Aside from CNNs, we craft an early-termination algorithm to improve efficiency of whole proposed approach further.
Apparently, homogeneous CUs have tendency to split into four sub-CUs and vice versa.
Considering that, we do a statistical analysis on every CU luma samples in our training sequence.
Given feasibility and computational complexity, we choose two most possible CU classification indicators, range and standard deviation of CU luma samples.
Every range and standard deviation samples of our training dataset are separated into two groups i.e. ones of non-split CUs and ones of split CUs.
Then, both selected indicators are assessed by probability density.

\begin{figure}[!h]
	\centering
	
	\subfigure[Range of $64\times64$ CUs]{
		\begin{minipage}{40mm}
			\includegraphics[width=42mm]{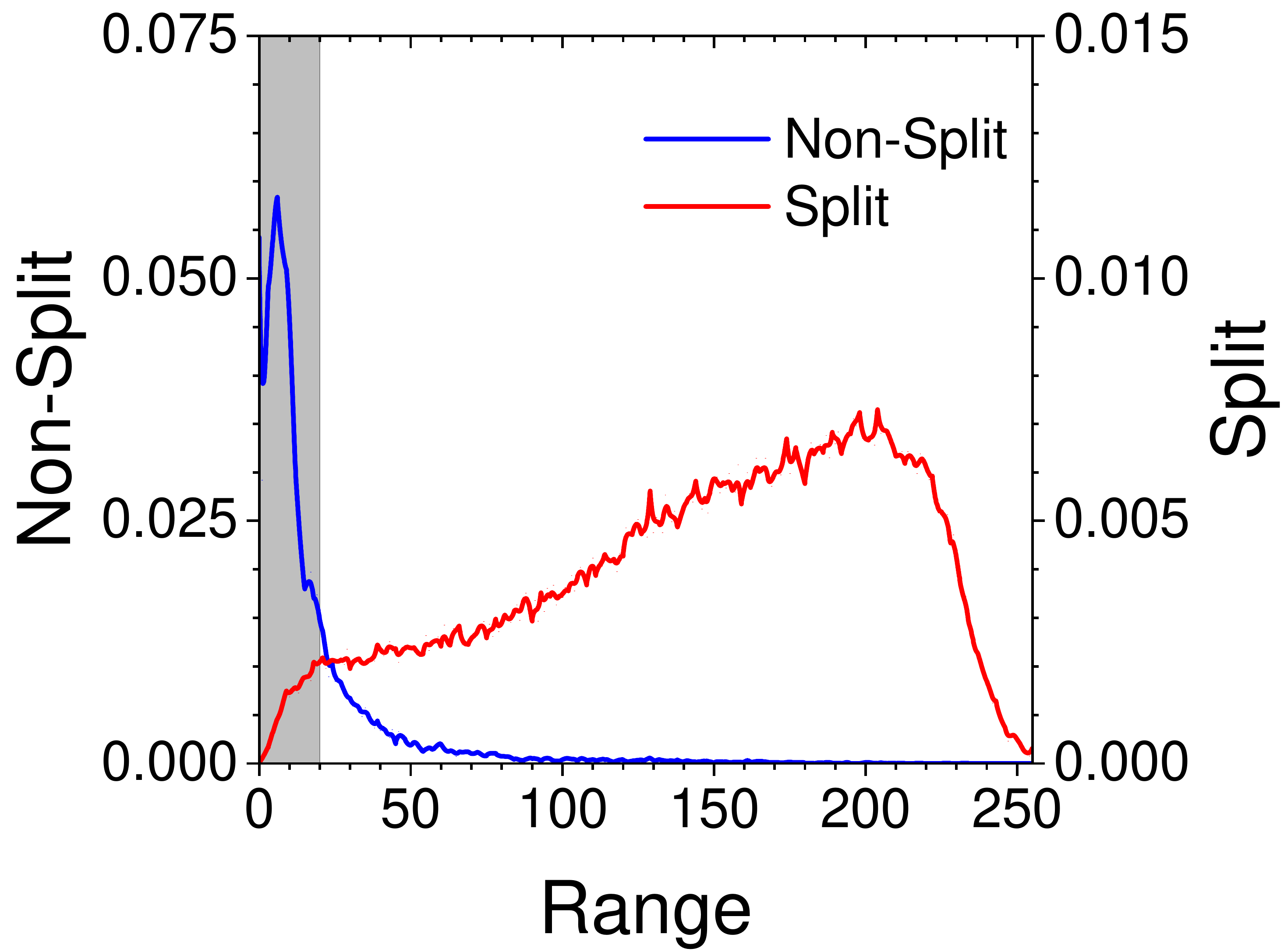}
			\label{fig:indicators:cu64_range}
		\end{minipage}
	}
	\subfigure[Std. dev. of $64\times64$ CUs]{
		\begin{minipage}{40mm}
			\includegraphics[width=42mm]{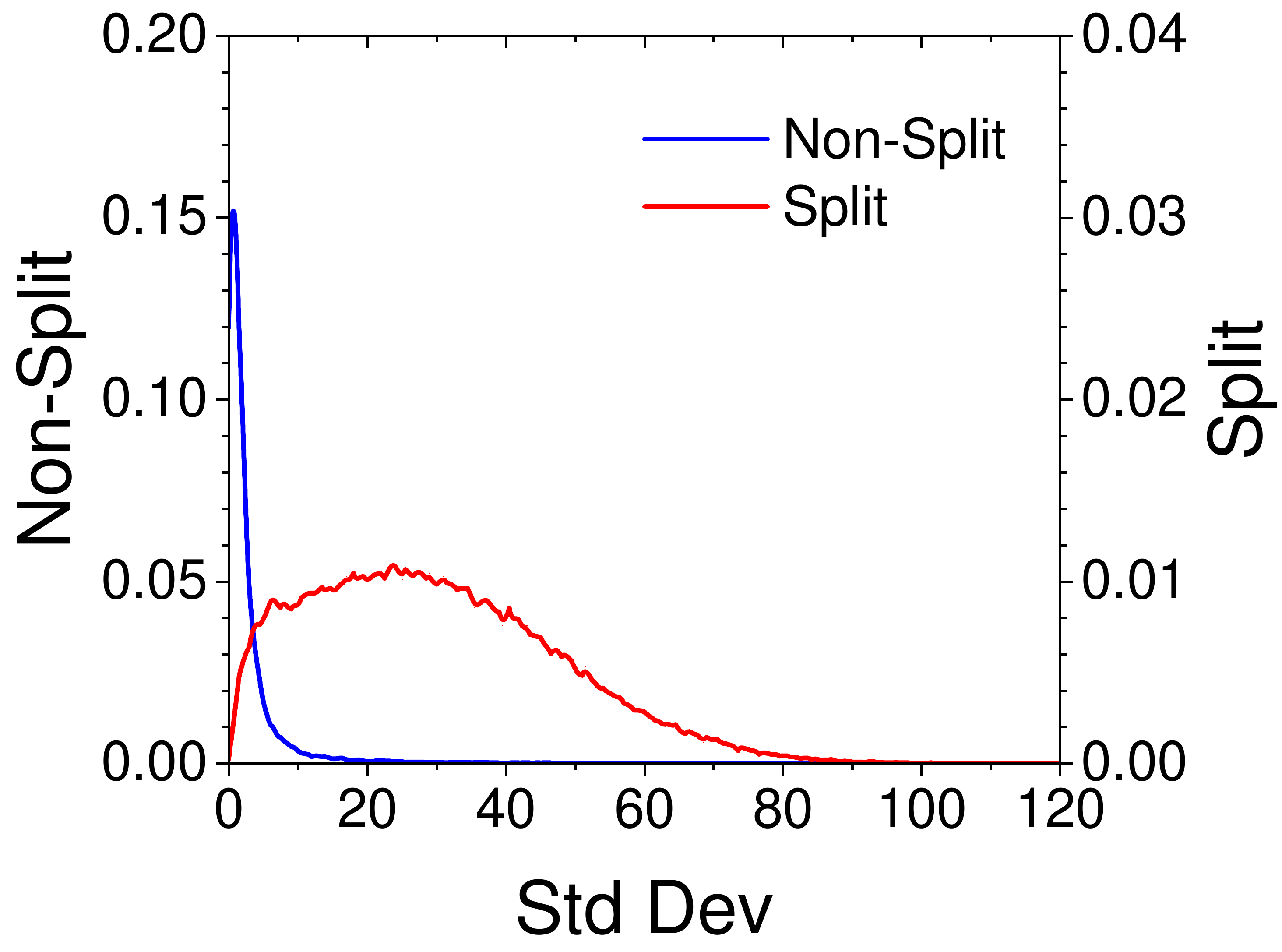}
			\label{fig:indicators:cu64_stddev}
		\end{minipage}
	}
	\subfigure[Range of $32\times32$ CUs]{
		\begin{minipage}{40mm}
			\includegraphics[width=42mm]{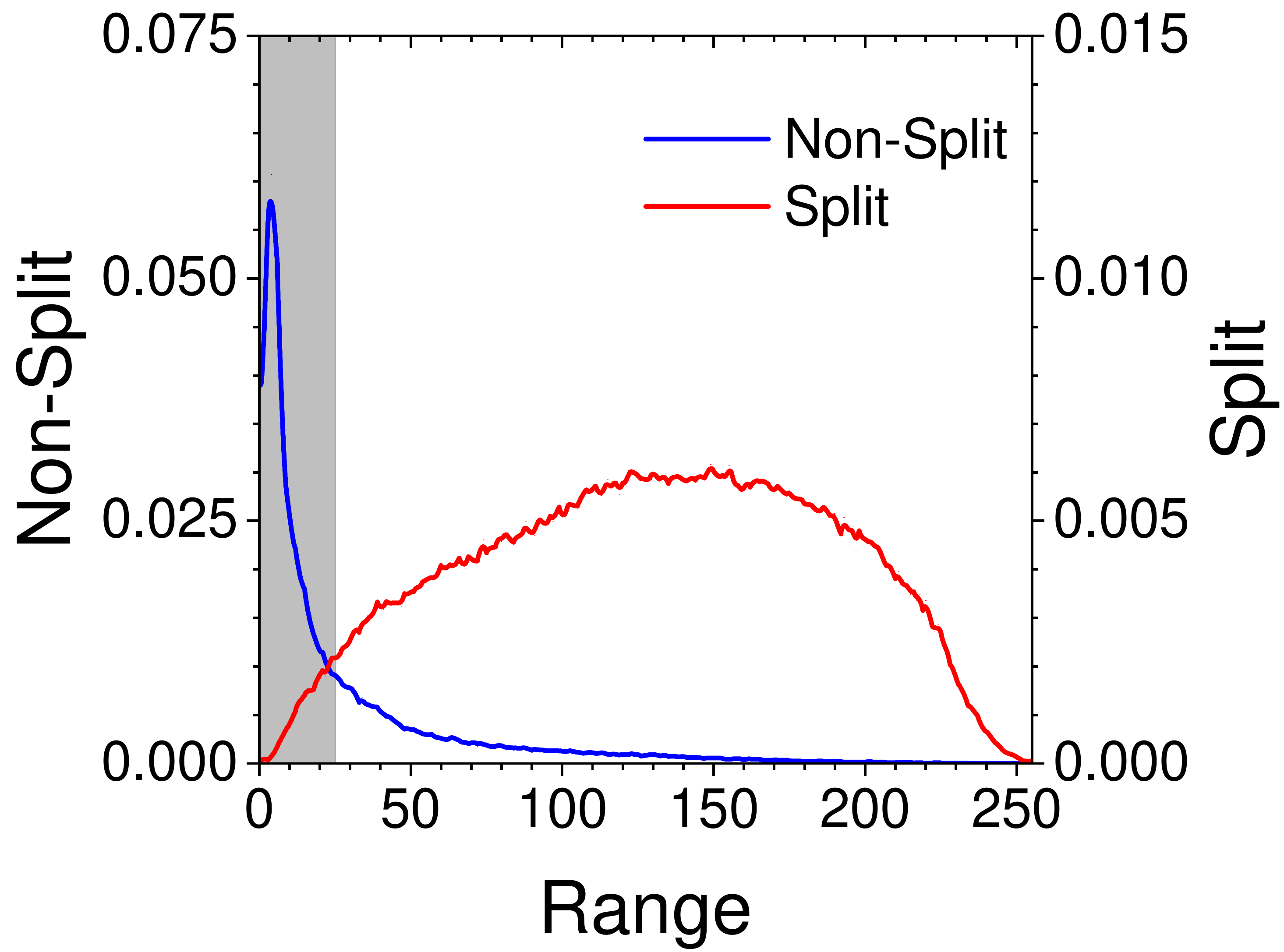}
			\label{fig:indicators:cu32_range}
		\end{minipage}
	}
	\subfigure[Std. dev. of $32\times32$ CUs]{
		\begin{minipage}{40mm}
			\includegraphics[width=42mm]{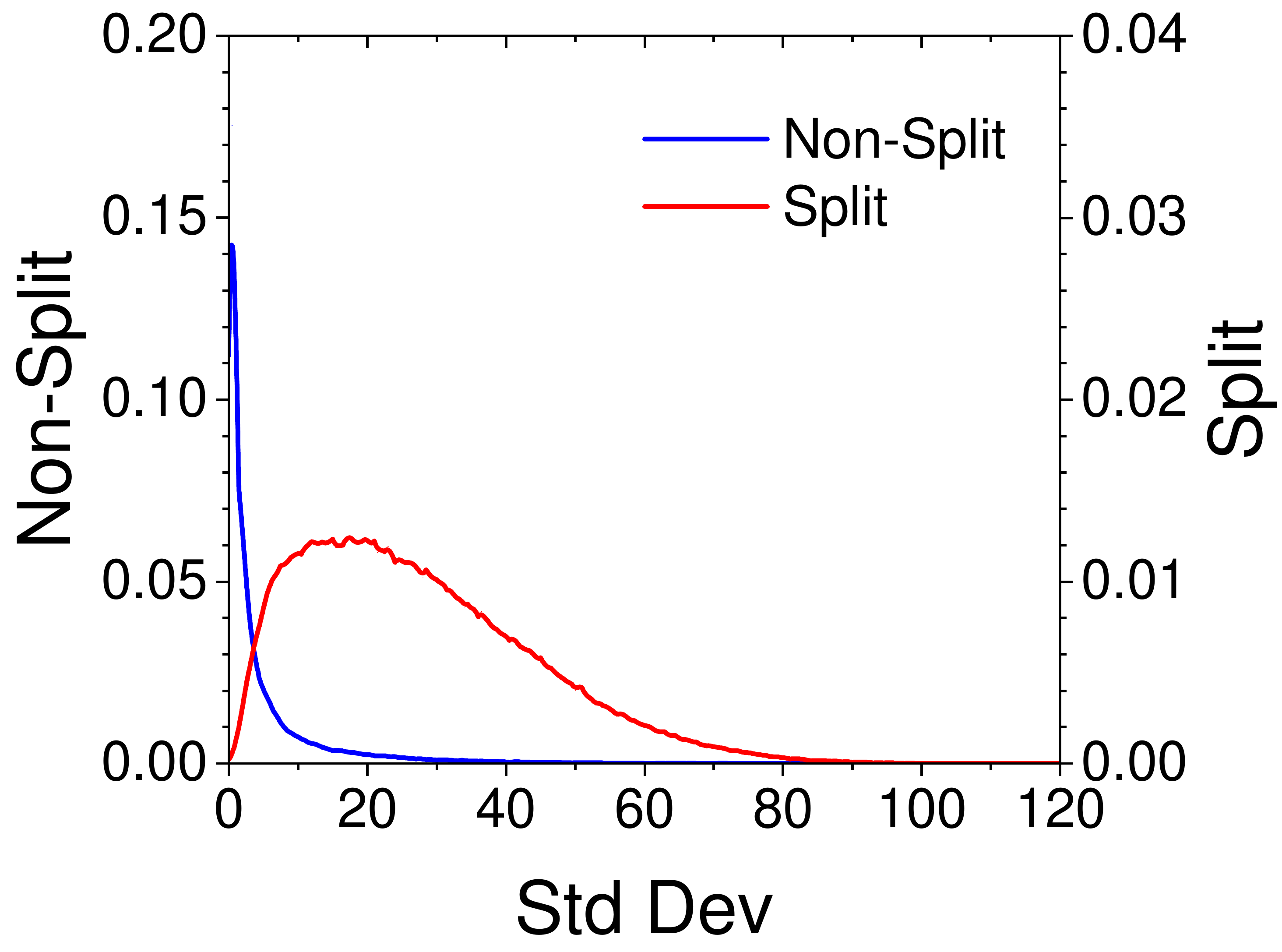}
			\label{fig:indicators:cu32_stddev}
		\end{minipage}
	}
	\subfigure[Range of $16\times16$ CUs]{
		\begin{minipage}{40mm}
			\includegraphics[width=42mm]{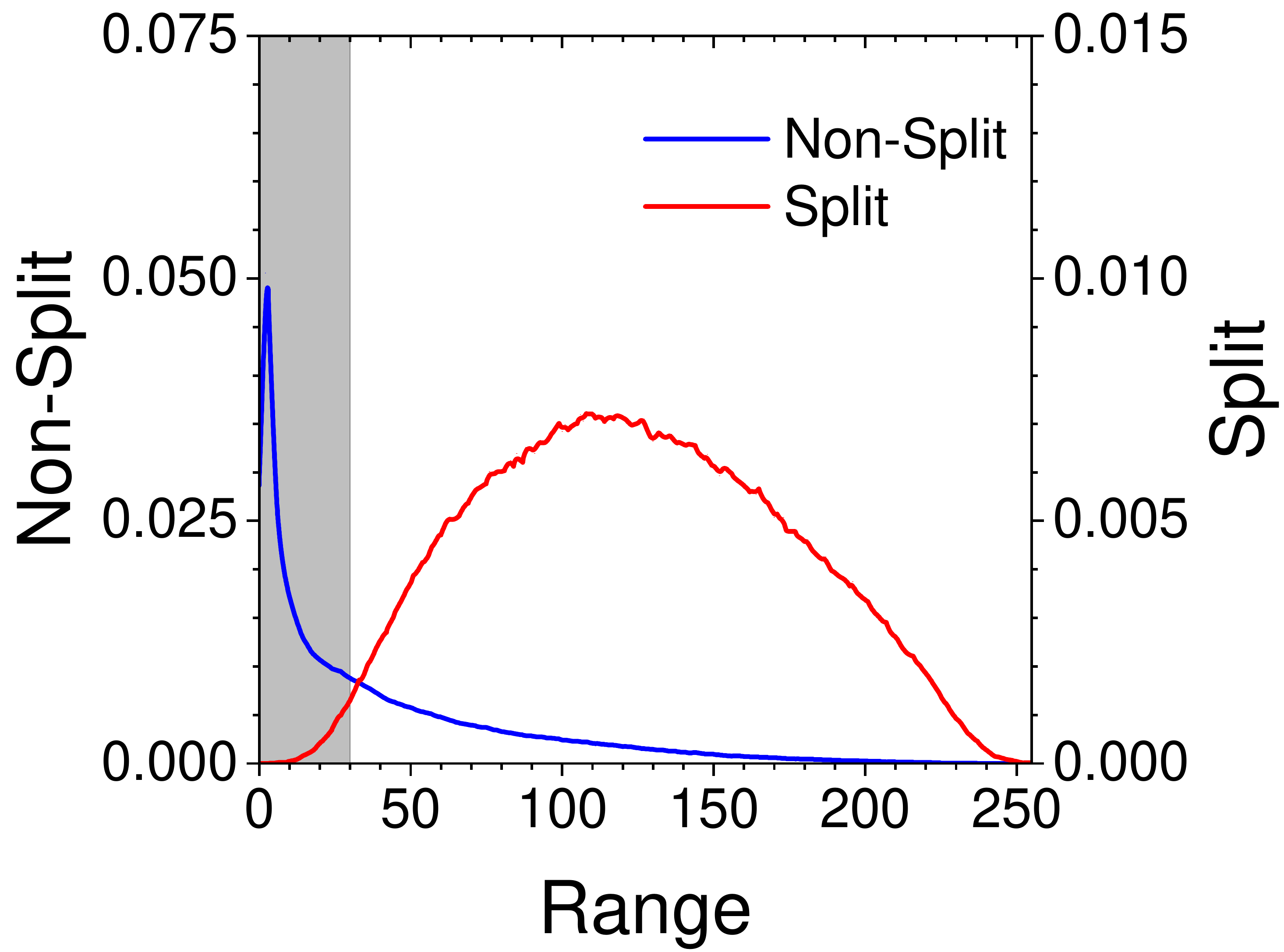}
			\label{fig:indicators:cu16_range}
		\end{minipage}
	}
	\subfigure[Std. dev. of $16\times16$ CUs]{
		\begin{minipage}{40mm}
			\includegraphics[width=42mm]{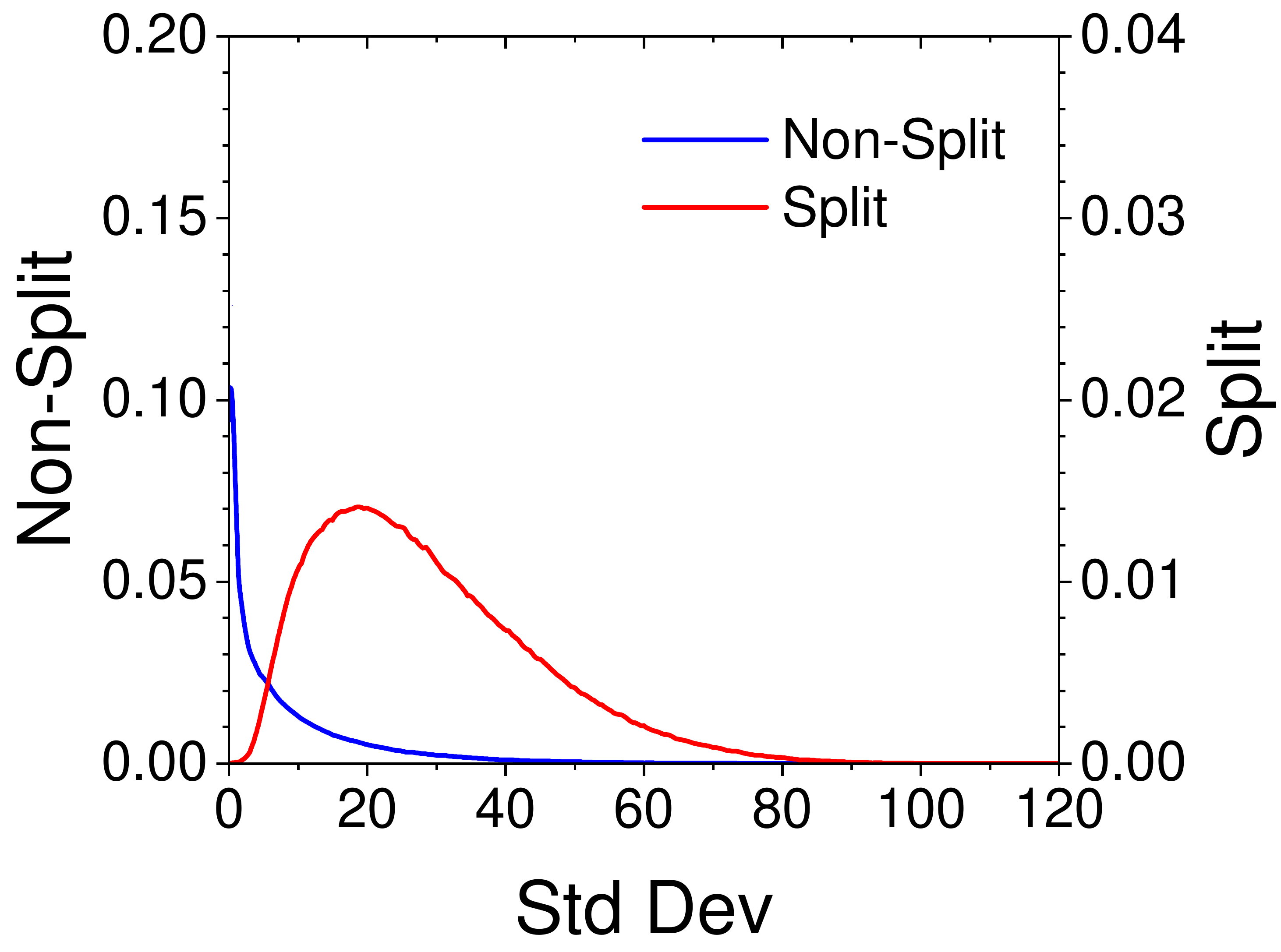}
			\label{fig:indicators:cu16_stddev}
		\end{minipage}
	}

	\caption{Probability density of both selected indicators}
	\label{fig:indicators}
\end{figure}

As Fig. \ref{fig:indicators} shows, the possibility densities of two selected indicators, range and standard deviation of CU luma samples, are shown respectively in each graph within non-split and split groups of three CU sizes.
As we can see, split CUs basically have larger range and standard deviation than non-split CUs, which results in an truth that CUs can be classified into non-split and split by their range and standard deviation.
However, the differential of range is obviously larger than standard deviation, as shown in Fig. \ref{fig:indicators}, which manifests that the range is more effective to distinguish non-split/split CUs.
Another advantage of range is that it is much easier and faster to calculate range than standard deviation.

Consequently, we regard the range of CU luma samples as the indicator to classify CU partition.
More clearly, the threshold are configured as 20 in $64\times64$ CUs, 25 in $32\times32$ CUs and 30 in $16\times16$ CUs.
CUs whose range of luma samples below the threshold will be early terminated in this stage as shown in grey areas in Fig. \ref{fig:indicators}.

\subsection{CNN-based CU Partition Algorithm}

In pursuit of higher accuracy than current CNN-based CU partition approaches, a deeper and wider CNN structure is introduced and crafted for CU partition decision in HEVC.
Nevertheless, upon the trade-off of accuracy and computation complexity, the computational complexity of CNNs has to be controlled in an acceptable limitation.
Hence we cannot only seek the highest accuracy by sacrifice of computational performance and the proposed CNNs are illustrated below.

$64\times64$, $32\times32$ and $16\times16$ CU will be determined to non-split or split during encoding process.
Thus, we design three large-scale CNNs to decide CU partition mode at each depth from 0 ($64\times64$) to 3 ($16\times16$).
In each CNN, the input layer is two-dimensional normalized luma samples of the CU, and the output layer is two softmax activation units representing the probabilities of non-split and split.
Our large-scale CNN structures are enlightened by GoogLeNet \cite{szegedy2015going} comprising several Inception layers.

\begin{figure*}[ht]
	\centering
	\subfigure[CNN for $64\times64$ CUs]{
		\includegraphics[width=175mm]{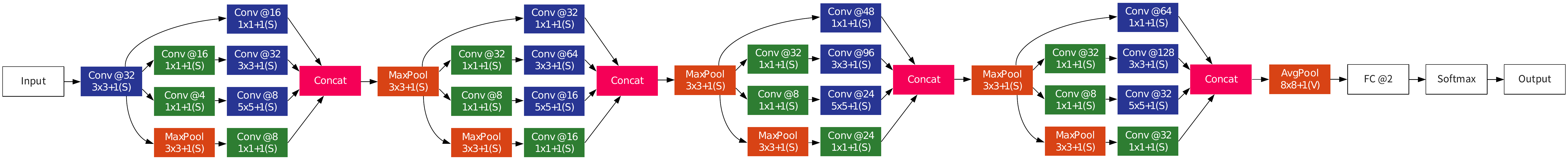}
		\label{fig:cnn:cu64}
	}
	
	\subfigure[CNN for $32\times32$ CUs]{
		\includegraphics[width=175mm]{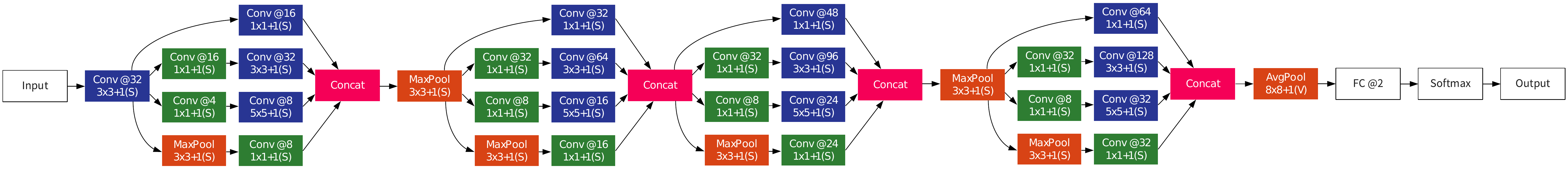}
		\label{fig:cnn:cu32}
	}
	
	\subfigure[CNN for $16\times16$ CUs]{
		\includegraphics[width=175mm]{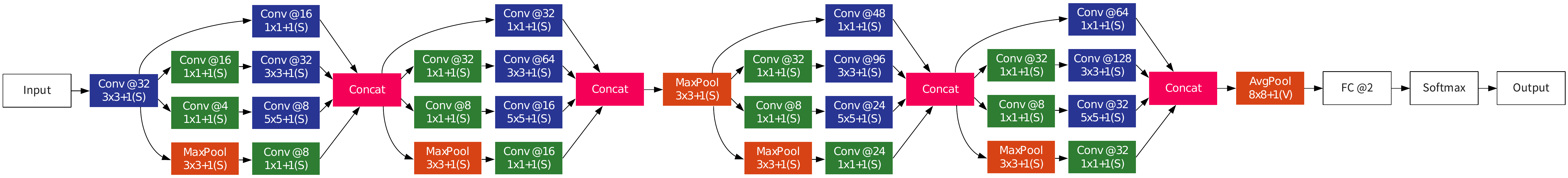}
		\label{fig:cnn:cu16}
	}
	\caption{Proposed CNN structures \protect\footnotemark[1]}
	\label{fig:cnn}
\end{figure*}

Particularly, frameworks and parameters of the proposed CNNs for each size of CU are elaborated in Fig. \ref{fig:cnn}.
At very beginning of each CNN, there is a $3\times3$ convolutional layer to increase input channel from 1 to 32.
Then, core framework of each CNN consists of four Inception layers for features extraction and many max pooling layer for dimension reduction.
In core framework, only sequence arrangement of four Inception layers and pooling layers is different from each other CNN, and four Inception layers in each CNN have same numbers of channels.
Specifically, max poling layers are followed every Inception layer for $64\times64$ CUs, the first and third Inception layer for $32\times32$ CUs, and the second Inception layer for $16\times16$ CUs.
For each CNN, the output of core framework is a $8\times8\times256$ feature map, and there is a $8\times8$ valid average pooling layer to flat that into a one-dimensional array.
Final layer is a fully connected layer and a softmax activation.

\subsection{Self-Adaptive Weighted Loss Function}

Our training dataset is extracted from original HEVC encoder, consisting of the luma samples (as feature) and the split flag (as label) of every CU in the training sequence.
To prevent overlap the test sequences, the training sequence is concatenation of 1,000 raw images that are arbitrarily selected from RAISE database \cite{dang2015raise}.
It should be noticed that as a matter of fact larger CUs tend to split, so our training dataset actually is a classification-imbalance dataset in which the ratio of positive and negative samples is far larger or smaller than 1 as shown in Table \ref{tab:dataset}.
It undoubtedly results in a severe problem in the training process.
That is under any circumstances CNNs, without significantly increasing loss, will always choose the partition mode of the majority of training samples.

\footnotetext[1]{Each block indicated a layer of CNN in the form of ``$Layer Type\ [@Channel] \ KernelWidth \times KernelHeight+Stride(Padding(S/V))$"}
\renewcommand\arraystretch{1.2}
\begin{table}[!h]
	\centering
	\caption{Partition Distribution of CUs in Training Dataset}
	\label{tab:dataset}

	\begin{small}
	\begin{tabular}{|p{16mm}<{\centering}
					|p{15mm}<{\centering}|p{15mm}<{\centering}|p{15mm}<{\centering}|}
		\hline
		{\bfseries CU Size}    & $64\times64$ & $32\times32$ & $16\times16$ \\
		\hline
		{\bfseries Non-Split}  &      508,105 &    3,828,038 &   29,467,909 \\
		\hline
		{\bfseries Split}      &    1,667,895 &    4,875,962 &    5,348,091 \\
		\hline
		{\bfseries NS/S Ratio} &         0.30 &         0.78 &         5.51 \\
		\hline
		{\bfseries Total}      &    2,176,000 &    8,704,000 &   34,816,000 \\
		\hline
	\end{tabular}
	\end{small}
\end{table}

Against this classification-imbalance problem, we propose the self-adaptive weighted loss function.
Based on the traditional cross-entropy loss function, our self-adaptive weighted loss function is given by
\begin{equation}
	\label{eq:loss}
	{L =  - \frac{1}{N}\sum\limits_i {\alpha_1 y_i\ln {{\hat y}_i} + \alpha_0 (1 - y_i)} \ln (1 - {\hat y_i})}
\end{equation}
where $\alpha_0$ and $\alpha_1$ is weight of 0 (non-split) samples' loss and 1 (split) samples' loss respectively and $N$ is batch size.
Parameter $\alpha_0$ and $\alpha_1$ will be adjusted during training process to adapt to the training results.
In detail, for every $M$ steps, the next $\alpha_1 / \alpha_0$ will be set to equal the weighted sum of current $\alpha_1 / \alpha_0$ of proportion of amounts of 0 samples and 1 samples in test results of present CNN.
When updating $\alpha_0$ and $\alpha_1$, keep $\alpha_0 + \alpha_1 = 2$ in order to control loss in a reasonable range.
Hence the update method of loss weights is given by
\begin{equation}
	\label{eq:loss_update}
	{(\frac {\alpha_1} {\alpha_0})_{t + 1} = (1 - \eta )(\frac {\alpha_1} {\alpha_0})_t + \eta \frac{N_0}{N_1}}
\end{equation}
where $N_0$, $N_1$ is the amounts of 0 samples and 1 samples in test results, and $\eta$ is a hyperparameter representing the weight of the proportion of two samples to update loss weights.
Since loss of minority of predictions will be multiplied a relatively large penalty item, the estimations of CNN will not remain the majority.
Note that $\alpha_0$ and $\alpha_1$ are a self-adaptive auxiliary parameters instead of hyperparameters.

\section{Experimental Results}

In our experiment, CPU is Intel Xeon CPU E5-2620 v4, GPU is NVIDIA GeForce GTX 1080 and OS is Ubuntu 16.04.
Our CNNs are implemented with TensorFlow \cite{abadi2016tensorflow} deep learning library
The proposed approach is integrated with HEVC Test Model (HM) 16.18, and we use the test sequences with QP = \{22, 27, 32, 37\} under all intra (AI) configuration.

\renewcommand\arraystretch{1.2}
\begin{table*}[ht]
	\centering
	\caption{BD-Bitrate(\%), BD-PSNR(dB) and Time Saving(\%) of the Proposed and Relative Approach}
	\label{tab:results}
	
	\begin{small}
	\begin{tabular}{|p{24mm}<{\centering}|p{18mm}<{\centering}
					|p{10mm}<{\centering}|p{10mm}<{\centering}|p{10mm}<{\centering}
					|p{10mm}<{\centering}|p{10mm}<{\centering}|p{10mm}<{\centering}
					|p{10mm}<{\centering}|p{10mm}<{\centering}|p{10mm}<{\centering}|}
		\hline
		\multirow{3}*{\bfseries Sequences} &
		\multirow{3}*{\bfseries Resolution} &
		\multicolumn{3}{c|}{\bfseries Liu \textit{et al.} \cite{liu2016cnn}} &
		\multicolumn{3}{c|}{\bfseries Cen \textit{et al.} \cite{cen2015fast}} &
		\multicolumn{3}{c|}{\bfseries Proposed} \\
		\cline{3-11}
		 & & {\bfseries BD-} & {\bfseries BD-} & \multirow{2}*{\bfseries TS} &
		 {\bfseries BD-} & {\bfseries BD-} & \multirow{2}*{\bfseries TS} &
		 {\bfseries BD-} & {\bfseries BD-} & \multirow{2}*{\bfseries TS} \\
		 & & {\bfseries BR} & {\bfseries PSNR} & &
		 {\bfseries BR} & {\bfseries PSNR} & &
		 {\bfseries BR} & {\bfseries PSNR} & \\
		\hline
		BQTerrace           &  $1920\times1080$ &            4.21  &            -0.10  &            36.71  &
												  {\bfseries 2.26} & {\bfseries -0.07} &            13.14  &
												             2.42  &            -0.14  & {\bfseries 43.30} \\
		\hline
		BasketballDrill     &    $832\times480$ &            5.67  &            -0.21  &            34.75  &
												             7.33  & {\bfseries -0.10} &            18.35  &
												  {\bfseries 2.36} & {\bfseries -0.10} & {\bfseries 37.71} \\
		\hline
		PartyScene          &    $832\times480$ &            4.34  &            -0.12  & {\bfseries 34.83} &
												             4.00  &            -0.18  &            11.45  &
												  {\bfseries 1.39} & {\bfseries -0.10} &            31.69  \\
		\hline
		RaceHorses          &    $416\times240$ &            5.23  &            -0.23  & {\bfseries 32.70} &
												  {\bfseries 0.01} &            -0.14  &            12.75  &
												             1.99  & {\bfseries -0.13} &            27.03  \\
		\hline
		BasketballDrillText &    $832\times480$ &            5.96  &            -0.26  & {\bfseries 37.66} &
												  {\bfseries 0.09} &            -1.82  &            18.90  &
												             4.42  & {\bfseries -0.22} &            35.52  \\
		\hline
		ChinaSpeed          &   $1024\times768$ &            4.66  &            -0.39  &            34.15  &
												  {\bfseries 1.14} & {\bfseries -0.11} &            20.02  &
												             4.11  &            -0.38  & {\bfseries 43.05} \\
		\hline
		SlideEditing        &   $1280\times720$ &            3.52  &            -0.50  &            31.69  &
												             2.42  & {\bfseries -0.13} &            13.70  &
												  {\bfseries 2.18} &            -0.34  & {\bfseries 38.84} \\
		\hline
		\multicolumn{2}{|c|}{\bfseries Average} &            4.79  &            -0.25  &            34.64  &
												             2.85  &            -0.36  &            15.47  &
												  {\bfseries 2.69} & {\bfseries -0.22} & {\bfseries 36.74} \\
		\hline
	\end{tabular}
	\end{small}
\end{table*}

We compare our proposed approach with other relative approaches by Liu \textit{et al.} \cite{liu2016cnn} and Cen \textit{et al.} \cite{cen2015fast} on our aforementioned experimental environment.
Cen's approach initially calculates the rarest depth (0 or 3) which will be skipped in subsequent algorithm, and then CU depth is determined from the remaining depths by the comparison algorithm with adjacent CUs.
The experimental results are shown in Table \ref{tab:results}, including the Bjontegaard delta bitrate(BD-BR), Bjontegaard delta peak signal-to-noise(BD-PSNR) \cite{bjontegaard2001calculation} and the encoding time saving (TS) of three approaches with original HM encoder as anchor.
And the time saving is given by
\begin{equation}
\label{eq:time_saving}
{TS = \frac {T_{orig} - T_{prop}} {T_{orig}} \times 100\%}
\end{equation}
where $T_{orig}$ and $T_{prop}$ is the encoding time of original and proposed encoder respectively.

Firstly, the proposed approach reduce 36.74\% computational complexity on average, and specifically at maximum 43.40\% in ``\textit{BQTerrace}".
As the brute-force RDO quad-tree search in original HM is skipped, the proposed approach is able to reduce such computational complexity.
Because the calculation of range of CU luma samples is fast and easy, our novel early-termination algorithm also assist to mitigate compuational complexity further.
Compared to the approach by Cen \textit{et al.} \cite{cen2015fast}, their approach reduces only 15.47\% encoding time and the proposed approach saves much more encoding time.

In addition, the proposed approach reach a better RD performance with 2.69\% BD-BR increment and 0.22dB BD-PSNR decrease.
As our two-stage approach has considerable accuracy on CU partition mode prediction, so that RD performance can not be aggravated.
The approach by Cen \textit{et al.} \cite{cen2015fast} has 2.85\% BD-BR increment which is higher than our approach.

Moreover, the shallow and narrow CNN-based approach by Liu \textit{et al.} \cite{liu2016cnn} on dedicated hardware incurs 4.79\% BD-BR rise.
However, our proposed approach achieves lower 2.69\% BD-BR rise.
Thus our large-scale CNN-based approach has stronger ability to extract and generalize CU features, but with relatively more computation burden on general hardware.
And the simple early-termination algorithm based on the range of CU luma samples mitigates this issue without accuracy loss.
Higher accuracy of CU partition estimation of the proposed two-stage approach makes great contribution to lower BD-BR rise.

Therefore, experimental results indicate that the proposed approach on CU partition acceleration outperforms conventional method in HM 16.18 with 36.74\% encoding time saving and negligible BD-BR increment.
In terms of both RD performance and encoding time saving the proposed approach is better than other approaches in \cite{liu2016cnn} and \cite{cen2015fast}.

\section{Conclusion}

In this paper, we proposed a novel CU partition decision approach in HEVC, which has been proved to have ability to reduce encoding time without significantly impairing RD performance.
As presented above, the proposed approach has two stages in which range of CU luma samples and three CNNs targeting each CU size are utilized to decide a CU partition mode.
Relieving computational complexity of CU partition decision, the proposed approach reduces nearly 37\% encoding time with ignorable 2.69\% BD-BR rise in comparison with the original HM 16.18 encoder.

% the following command shrinks the final page to force the columns to
% be balanced.  You will need to adjust the value according to the
% appearance of your last page.  Start by setting the value to 0mm
% and slowly increase it until the columns balance.  Alternatively,
% use balance.sty to do the job.
\enlargethispage{0mm}

\bibliographystyle{IEEEtran}

\bibliography{IEEE-conference-A4-LaTeX-format}

\end{document}